\documentclass{article}

\usepackage{pslatex}
\usepackage{apacite}
\usepackage{graphicx}
\usepackage{amsfonts} 
\usepackage{amsmath}
\graphicspath{ {./images/} }
\usepackage[font=small,skip=0pt]{caption}
\usepackage{geometry}
\title{Characterising representation dynamics in\\ recurrent neural networks for object recognition}
 
\author{{\large \bf Sushrut Thorat \qquad Adrien Doerig \qquad Tim C. Kietzmann} \\
sushrut.thorat94@gmail.com, adoerig@uos.de, tkietzma@uos.de\\
  Institute of Cognitive Science, Osnabrück University, Germany}

\begin{document}

\maketitle

\section*{Abstract}
{
\bf
Recurrent neural networks (RNNs) have yielded promising results for both recognizing objects in challenging conditions and modeling aspects of primate vision. However, the representational dynamics of recurrent computations remain poorly understood, especially in large-scale visual models. Here, we studied such dynamics in RNNs trained for object classification on MiniEcoset, a novel subset of ecoset. We report two main insights. First, upon inference, representations continued to evolve after correct classification, suggesting a lack of the notion of being ``done with classification''. Second, focusing on ``readout zones'' as a way to characterize the activation trajectories, we observe that misclassified representations exhibit activation patterns with lower L2 norm, and are positioned more peripherally in the readout zones. Such arrangements help the misclassified representations move into the correct zones as time progresses. Our findings generalize to networks with lateral and top-down connections, and include both additive and multiplicative interactions with the bottom-up sweep. The results therefore contribute to a general understanding of RNN dynamics in naturalistic tasks. We hope that the analysis framework will aid future investigations of other types of RNNs, including understanding of representational dynamics in primate vision\footnote{This article is a revision of the 2023 Conference on Cognitive Computational Neuroscience (CCN) paper, in which we present a new analysis in the Appendix, and include suggestions made by the CCN reviewers.}.
}
\begin{quote}
\small
\textbf{Keywords:} 
recurrent neural networks, object recognition, neural representations, dynamics, naturalistic tasks, readout zones
\end{quote}

\section{Introduction}

Feedback connections are ubiquitous in brains~\cite{felleman1991distributed}. The resulting recurrent computations are advantageous in challenging conditions such as recognizing objects in clutter~\cite{wyatte2014early,kreiman2020beyond} and natural scenes~\cite{spoerer2020recurrent}. Research into the representation dynamics underlying recurrent computations is nascent but accelerating~\cite{mante2013context,zamir2017feedback,quax2018emergent,mastrogiuseppe2018linking,van2020going,thorat2021category,lindsay2022bio,driscoll2022flexible}. Moving to a more naturalistic setting, this work investigates representations and their dynamics in a deep recurrent convolutional neural network (RNN), as they contribute to improving classification responses to natural images. While we provide novel insights into temporal trajectories of the RNNs, the developed framework applies more broadly and can be applied to both, artificial and biological neural network dynamics, and hence contributes to the toolbox available to researchers interested in modelling vision with deep neural networks~\cite{doerig2023neuroconnectionist}.

\begin{figure}[t]
  \begin{center}
  \includegraphics[width=\textwidth]{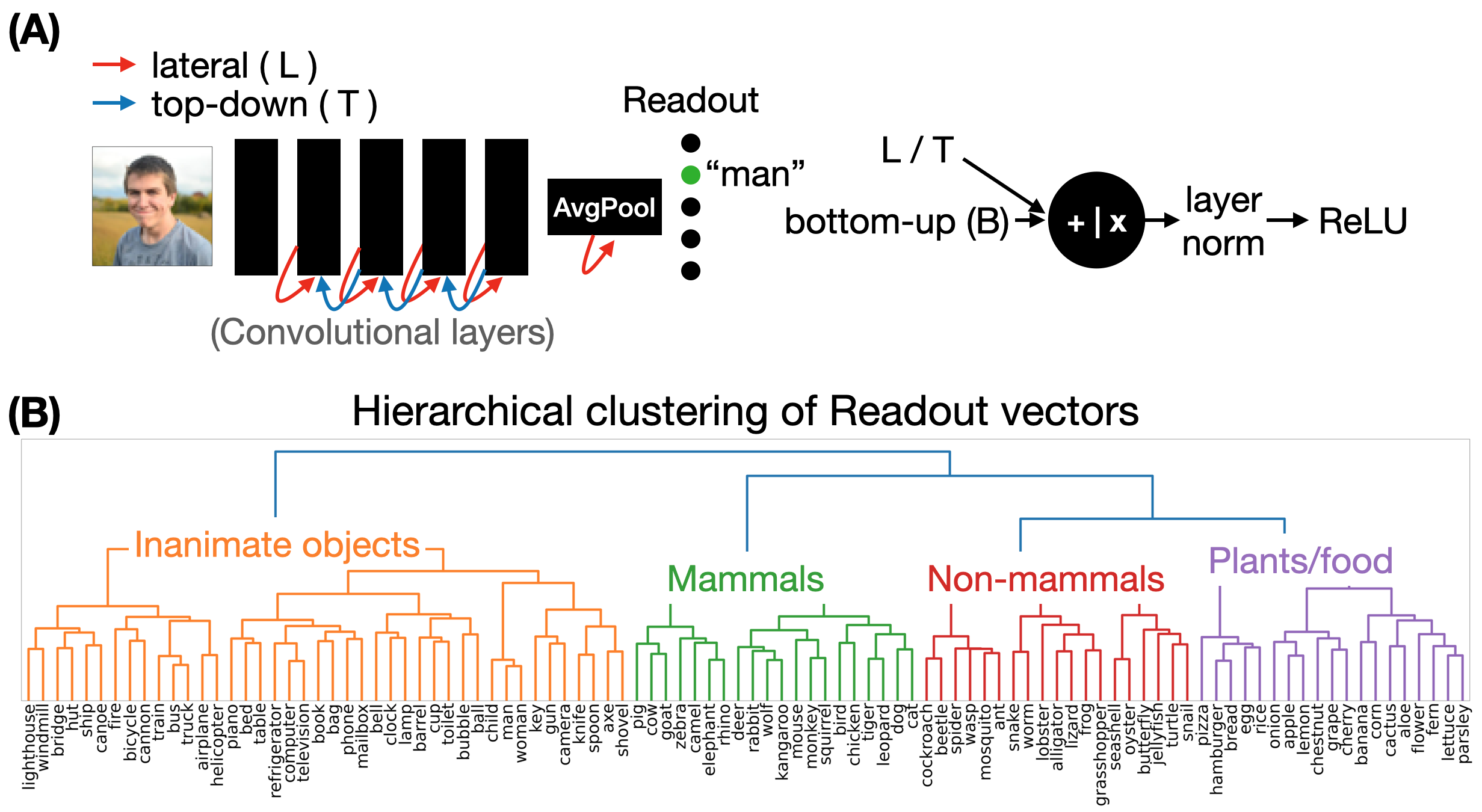}
  \end{center}
  \caption{(A) Architecture of the recurrent neural network. The lateral or top-down connections interact with bottom-up processing additively or multiplicatively. (B) Readout vectors capture referential semantic features of the data.}\label{fig1} 
  \end{figure}

\section{Model system and dataset}

In our RNN models\footnote{The training and evaluation scripts can be found at: \url{github.com/KietzmannLab/BLT-pytorch-CCN23}}, lateral or local top-down connections are included (Fig.~\ref{fig1}A). Such RNNs have been used as models of human neural dynamics and behavior~\cite{kietzmann2019recurrence,spoerer2020recurrent,doerig2022semantic}. The lateral and top-down connections interacted with the bottom-up sweep through either additive or multiplicative interactions. The RNNs were unrolled for $10$ timesteps. The RNNs were trained to classify the input images at each timestep (their readouts had no bias terms; see Appendix~\ref{ssc:rshapeimp}). The $64\times 64\,$px RGB images were taken from MiniEcoset\footnote{MiniEcoset can be found at: \url{osf.io/msna2/}}, which is a novel subset of ecoset~\cite{mehrer2021ecologically} containing $100$ object classes that follow a hierarchical object structure.

\section{Analysis}

We start our analyses by focusing on an RNN with lateral connections which interact with the feedforward sweep additively. Please note that these results generalize across RNN configuration (Fig.~\ref{fig3}A).

\subsection{Learned categorical structure}

We start our analysis by asking whether the RNN successfully learns the hierarchical structure encoded in the dataset statistics. To do so, we computed the similarities between the readout vectors (rows of the readout weight matrix, corresponding to connections from the final AvgPool layer to each of the readout neurons), as they can give us insight into which classes are considered similar by the RNN. Cosine similarity ($\vec{A}\cdot\vec{B}/|\vec{A}||\vec{B}|$) was computed between each pair of the readout vectors. 

Hierarchical clustering on the pairwise similarities revealed meaningful clusters (Fig.~\ref{fig1}B) resembling the dataset structure and the animacy organization observed in primate brains~\cite{grill2014functional}. This suggests that our choice of architecture and dataset leads to an interpretable feature extractor. 

\subsection{Convergent representation dynamics}

Next, we moved into analysing the representational dynamics of the RNNs, asking whether they exhibit a signature of being ``done with classification'', as expected in a stable RNN with attractor dynamics~\cite{linsley2020stable}. Additionally, we asked if the changes in pre-readout representations (i.e., final AvgPool layer activations) are smaller for images that are already correctly classified as opposed to images that are not yet correctly classified. For this analysis, we focused on images that were classified correctly and consistently starting from a given timestep $t$ (termed stable classification with $t_{stable} = t$; we only consider these images for subsequent analyses). To define representational changes, we analysed the $l^2$-norms of the change in representations across time, as a function of $t_{stable}$. 

\begin{figure}[t]
\begin{center}
\includegraphics[width=\textwidth]{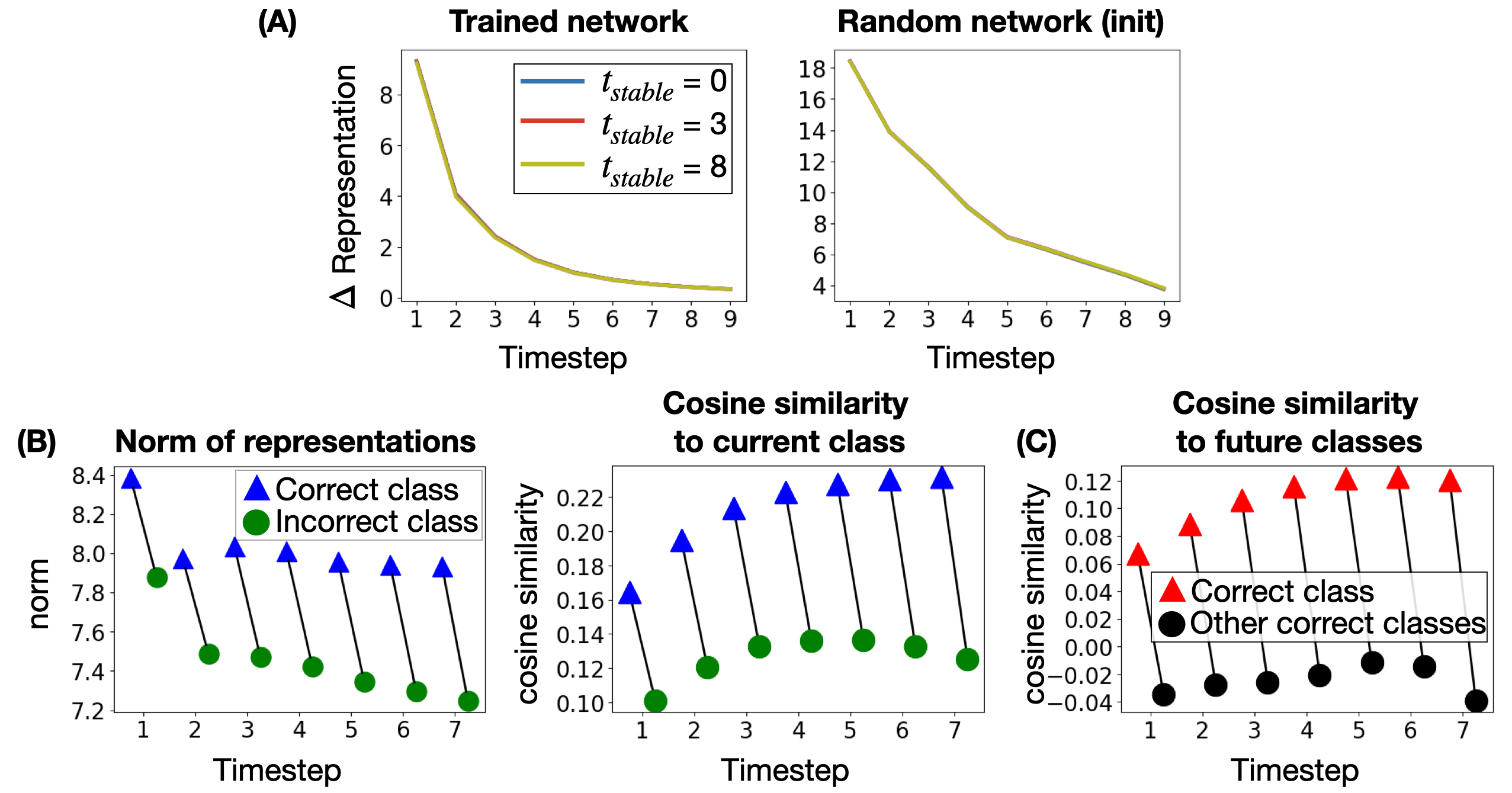}
\end{center}
\caption{(A) The amount of change in representations does not depend on the correctness of classification, in both trained and random RNNs. (B) Signatures of stable correct classification: the norm of the representation and its cosine similarity to the readout vector of the current class are higher. (C) Signature of the future correct class: for currently misclassified representations, the cosine similarity to the correct class readout vector is higher.}\label{fig2} 
\end{figure}

As seen in Fig.~\ref{fig2}A (left), the amount of representational change did not depend on $t_{stable}$: the changes in representations were not smaller for images that were classified correctly at earlier timesteps. However, the change in all representations did decrease with timesteps. 
These results indicate that although all representations ``settle'' across time, the rate of settling is independent of the correctness of classification.
Interestingly, as seen in Fig.~\ref{fig2}A (right), this reduction in the rate of change was also observed pre-training, suggesting these dynamics are a property of the network architecture. 
Finally, note that in contrast to previous findings~\cite{linsley2020stable}, these RNNs exhibits stable state dynamics despite them being trained with backpropagation through time (BPTT), as discussed in Appendix~\ref{ssc:stability}.

\subsection{Signatures of stable classification}

Originating from the observation that, on average, representations move the same distance regardless of correct classification, we hypothesized that representations that are able to transition into another class may initially be closer to the decision boundary, whereas the ones that do not transition are initially far from the boundary (and are therefore unable to leave the current class). As we show in the Appendix~\ref{ssc:rshape}, in networks with linear readouts and argmax decisions, the ``readout zones'', in which representations are assigned to a given class, resemble conical structures (a 2D schematic is shown in Fig.~\ref{fig3}B). Given this structure, being closer to the decision boundary either entails having a lower L2 norm or having a lower cosine similarity with the readout vector (see Appendix~\ref{ssc:rshapeimp} for further explanation). To explore this hypothesis, we assessed whether currently incorrectly-classified representations (that will eventually become correct) indeed have lower norms and/or lower cosine similarities with the readout vector of the current class.

At each timestep $t$, we compared both properties of the representations with $t_{stable} \leq t$ (i.e., currently correct) and the representations with $t_{stable} > t$ (i.e., currently incorrect): their norms, and their cosine similarities to the current readout. 
As seen in Fig.~\ref{fig2}B, both properties were smaller for $t_{stable} < t$ than for $t_{stable} \leq t$: the norms and cosine similarities were lower for representations that were incorrectly classified at a given timestep. As seen in Fig.~\ref{fig3}A, these patterns (averaged across timesteps) are independent of the kind of feedback used or how it interacts with the bottom-up sweep. This confirms the hypothesis that currently incorrect representations are closer to the decision boundary.

What constrains incorrect images to be closer to the decision boundary? There are two main possibilities: either any feedforward sweep, including in a purely feedforward network, automatically projects them to this position, or the feedforward sweep is shaped by the fact that recurrent computations move representations the same distance regardless of correct classification. To answer this question, we tested if the norms of two feedforward networks could predict the how fast images are correctly classified by the RNN. They do, as can be seen in Appendix~\ref{ssc:ffnet}. This suggests that the requirement of the recurrent computations (representations moving out of a class should be closer to the decision boundary) are satisfied by the representations instantiated by the feedforward sweep. The reason why incorrect images are closer to the decision boundary is independent of recurrence, and the effect of recurrent computations is to move these representations from the incorrect to the correct class. What properties of the images lead to their representations being initializated closer to the decision boundary remains to be explored.

\subsection{Signatures of the correct class}

\begin{figure}[t]
\begin{center}
\includegraphics[width=\textwidth]{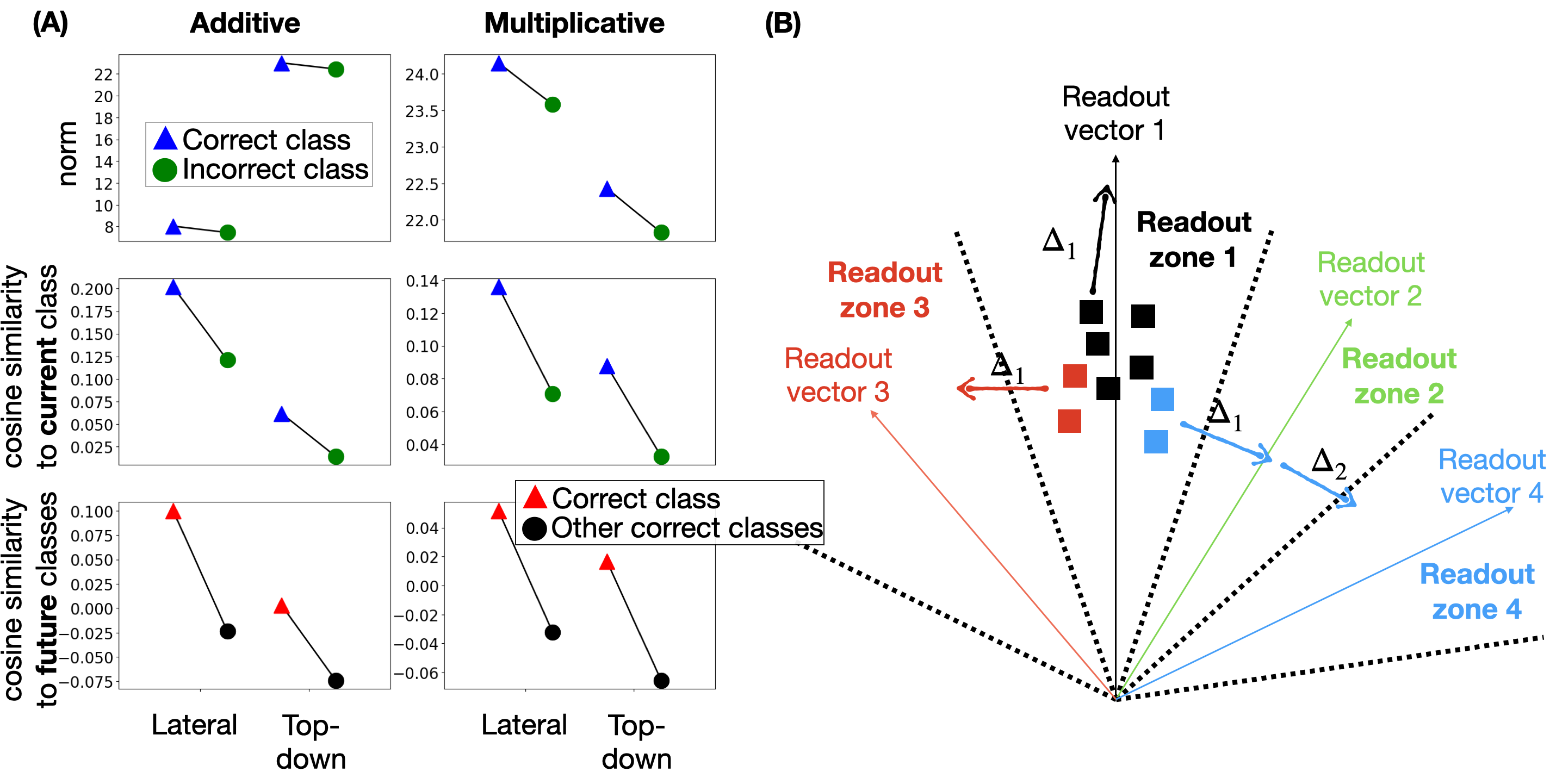}
\end{center}
\caption{(A) The results, shown in the previous figures, are independent of the type of feedback (lateral vs top-down) and its interaction with the bottom-up sweep (additive or multiplicative). (B) Schematic of the learned representation arrangements: incorrectly classified representations have a lower norm and lie closer to the class boundaries which allows for easier transition to another class.}\label{fig3} 
\end{figure}

We have now established that currently misclassified objects reside closer to the decision boundary (in the incorrect readout zone). Do these currently incorrectly-classified representations exhibit any signatures of their correct classes? Evidence for this would be provided if the cosine similarity of an incorrectly-classified representation to its correct class readout vector was higher than its cosine similarity to the readout vectors corresponding to the correct class of other incorrectly-classified representations in the same readout zone (see Fig.~\ref{fig3}B for a schematic). As seen in Fig.~\ref{fig2}C, the cosine similarity of the incorrectly-classified representations to the corresponding correct class readout vector is indeed higher than the cosine similarity to other correct classes' readout vectors. Hence, there are signatures of the correct classes in the incorrectly classified representations. This pattern (averaged across timesteps) is independent of the kind of feedback (lateral vs. top-down) and how it interacts with the bottom-up sweep (additive vs. multiplicative; Fig.~\ref{fig3}A). 

An intriguing question that arises from this is whether and how recurrent computations utilize these nascent features to correct the classification. Future work in understanding these dynamics shall consider: How do the incorrectly classified representations move through other classes to arrive at their correct classes? How do the feedback connections hierarchically (given Fig.~\ref{fig1}B) constrain the feedforward sweep to lead to those trajectories? Are similar dynamics/representations found in biological visual systems?

\section{Conclusions}

In the RNNs studied here, the magnitude of changes in network activations are surprisingly similar across images and decrease with model timesteps. This shows that the extent of recurrent dynamics experienced by image representations does not depend on the correctness of classification. In addition, we highlight an interesting representation arrangement, presented schematically in Fig.~\ref{fig3}B: image representations that are currently incorrectly classified (red and blue squares) have lower norms, and are closer to the current readout zone's decision boundary. The initial norm of the representation depends on the alignment of the image features with the feedforward weights, and can be seen as indicating the certainty of the network's inference after the feedforward sweep. For representations where certainty is low, recurrence can more easily move them towards the correct readout zone.

This work reported our first advances in deriving a framework for understanding representational dynamics in RNNs trained on naturalistic images, which we hope will further clarify how recurrent systems, both artificial and biological, reach their decisions. Future work should investigate the representation trajectories in other recurrent systems, including spatiotemporal data from the primate visual system.

\section{Acknowledgments}
The project was  partially funded by the European Union (ERC, TIME, Project $101039524$). Compute resources were funded by the Deutsche Forschungsgemeinschaft (DFG, German Research Foundation, Project number $456666331$).

\bibliographystyle{apacite}
\setlength{\bibleftmargin}{.125in}
\setlength{\bibindent}{-\bibleftmargin}

\bibliography{biblio}

\section{Appendix}

\subsection{The shape of the readout zones}
\label{ssc:rshape}

 Similar to most neural networks used for object recognition, in our RNNs, the readout layer consists of a linear transformation of the representation followed by the softmax operation. Given representation $r \in \mathbb{R}^{d_r}$, the readout weight matrix $M \in \mathbb{R}^{d_o\times d_r}$, and bias $b \in \mathbb{R}^{d_o}$, the softmaxed output responses $o \in \mathbb{R}^{d_o}$ are:

 \begin{equation}
  \begin{split}
   o' & = Mr+b \\
   o_i & = e^{o'_{i}} / \sum_{j=1}^{d_o} e^{o'_{j}}
  \end{split}
 \end{equation}

 As $e^x$ is a monotonic function, softmax does not change the rank-order of the responses: $\text{argsort}(o') = \text{argsort}(o)$. Therefore, the output of the network is the class represented by neuron $i$, which is associated with the readout vector $M[i,:]$ that has the largest dot product with $r$ (if bias $b=0$, which is the case in our RNNs). The rank-order of the dot products, $o'_i = \left \| M[i,:] \right \|\left \| r \right \|\text{cos}\theta$, is invariant to the L2 norm of the representation, $\left \| r \right \|$, with the other factors held constant. 

 \begin{figure}[t]
  \begin{center}
  \includegraphics[width=\textwidth]{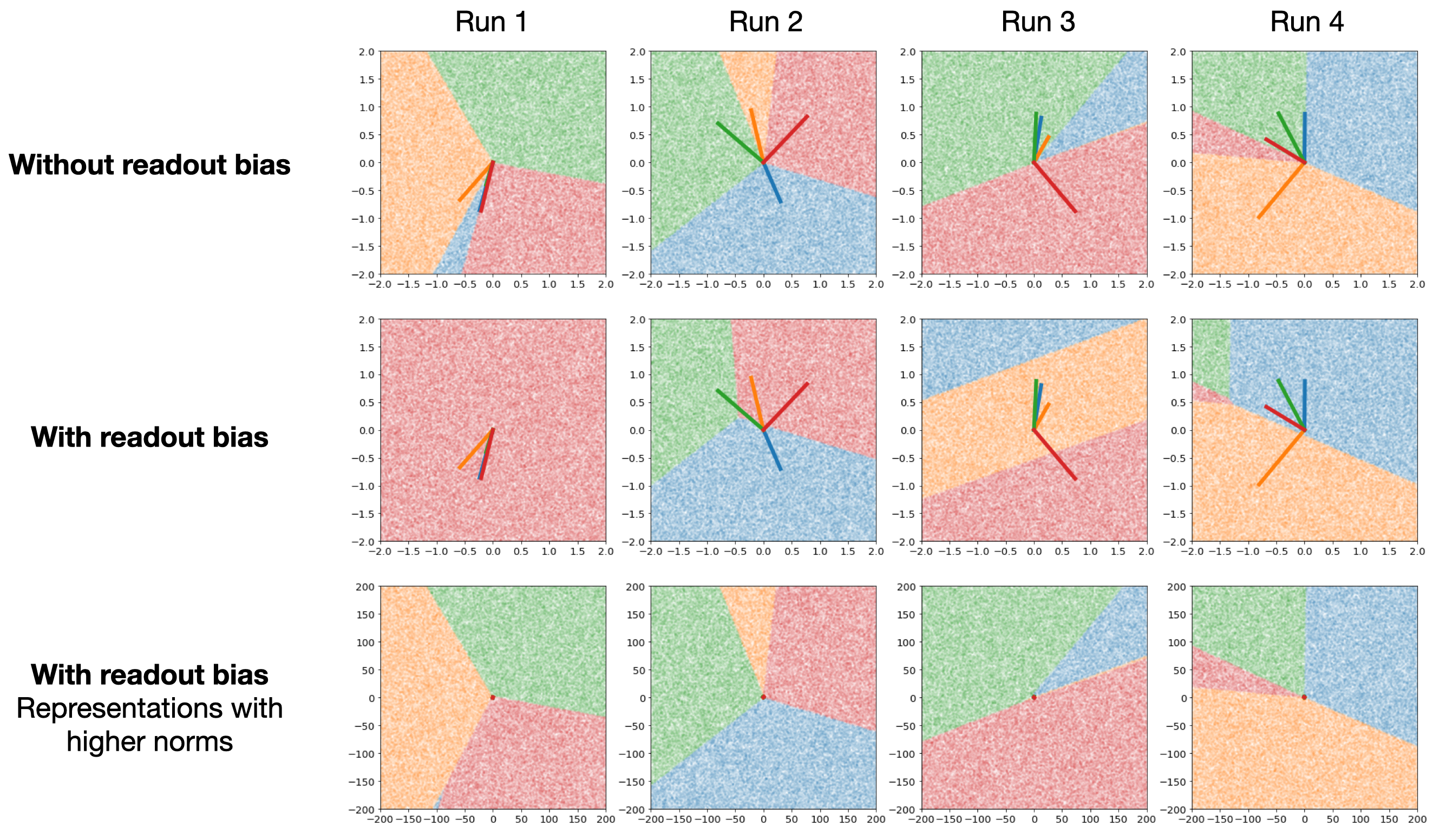}
  \end{center}
  \caption{Simulations of $4$ readout zones in a 2D representation space, with or without readout bias (the readout weights stay constant across a run). The corresponding readout vectors are also shown. Without readout bias, the zones resemble triangular areas with one vertex each at origin. Readout bias can add offsets to the zones, which are better visualized in the regime where the norm of the representations is much higher than the norm of the readout bias. Note that, in this low-dimensional readout regime, the readout vectors sometimes do not fall in the corresponding readout zone. However, with higher dimensional readouts, this does not happen (see Fig.~\ref{fig5})}\label{fig4} 
\end{figure}
 
 For $d_o=2$, this implies the regions of representation space corresponding to the network output decisions, ``readout zones'', would resemble triangles, with one of their vertices at origin. This is confirmed by the simulations of $4$ readout zones, in a 2D representation space subjected to varying random $4\times 2$ linear transformations and softmax (Fig.~\ref{fig4} (top)). The addition of readout bias corresponds to adding offsets to the readout zones as seen in Fig.~\ref{fig4} (middle). However, these readout zones still look like triangles in the regime where the norm of the representations is much higher than the norm of the readout bias (Fig.~\ref{fig4} (bottom)).

 \subsubsection{Implications for our results}
 \label{ssc:rshapeimp}

As mentioned earlier, our RNNs have no readout bias. Hence, the readout zones can be interpreted as higher-dimensional conical structures akin to the triangular regions in 2D, with one of their vertices at origin, supporting the norm-based representation arrangement in Fig.~\ref{fig3}B. However, Fig.~\ref{fig4} suggests that the readout vectors might not be inside the readout zones, which runs counter to our cosine similarity analysis. In Fig.~\ref{fig4}, the number of readout vectors is higher than the dimensionality of the representational space, whereas in our RNNs, the representational space is $512$D while the readout is $100$D. We checked if, in our setting, the readout vectors are indeed inside the readout zones.

First, we checked if the readout vectors are classified as the correct class, which would indicate they lie within their corresponding readout zones. If the readout matrix is $M \in \mathbb{R}^{100\times 512}$, the class decisions are given by: $\text{argmax}(M\times M^T,0)$. We found that $100\%$ of the readout vectors (from the RNN in Fig.~\ref{fig2}) were classified correctly i.e. they lay in their corresponding readout zones.

\begin{figure}[t]
  \begin{center}
  \includegraphics[width=0.65\textwidth]{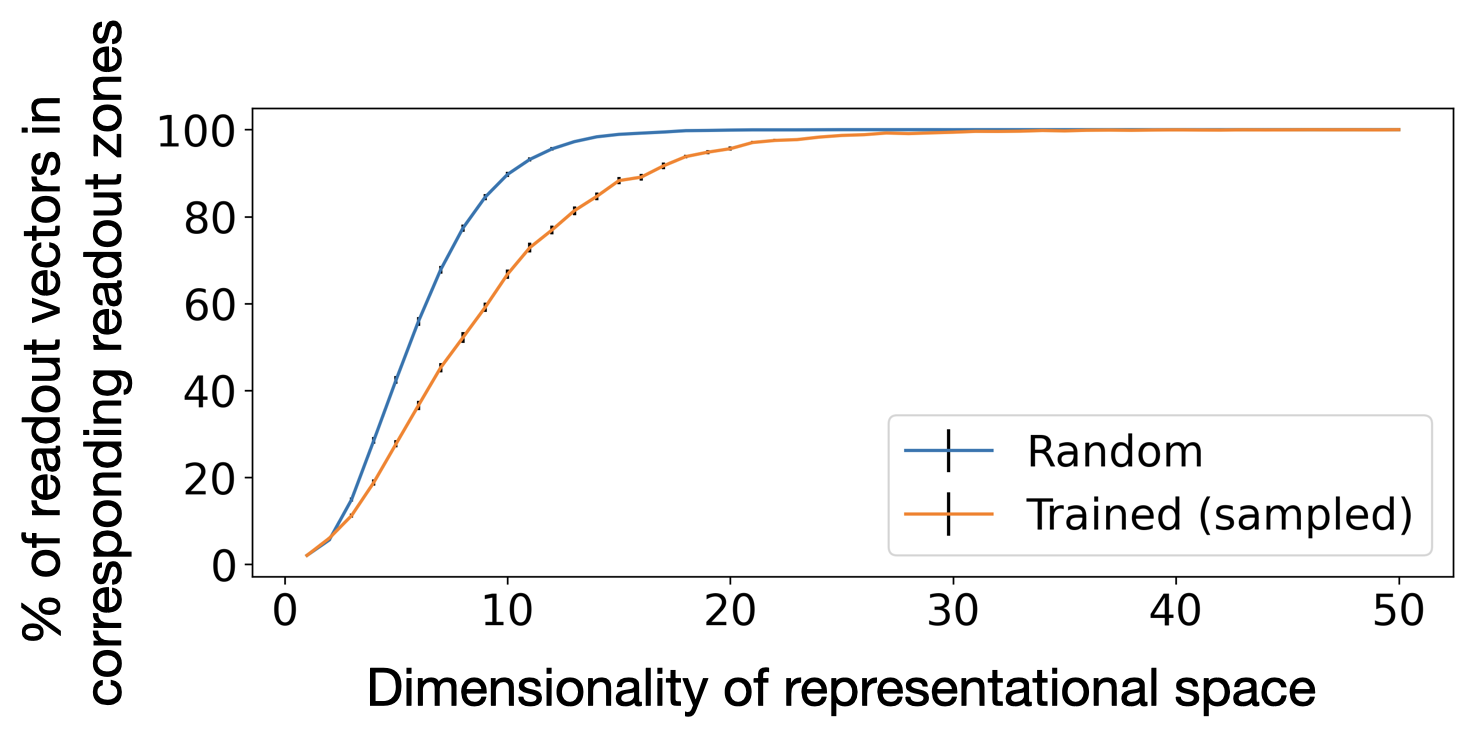}
  \end{center}
  \caption{The proportion of $100$ readout vectors laying in their corresponding readout zones increases and quickly reaches 100\% with the dimensionality of the representational space. Trained readouts are obtained from the readout matrix of the trained RNN (Fig.~\ref{fig2}), by sampling the number of representational dimensions as mentioned on the x-axis. $95\%$ confidence intervals of the mean of $100$ sampling repetitions are shown in black.}\label{fig5} 
\end{figure}

Second, we checked if the representations are in a high-dimensional space and the readouts span a subspace, the proportion of readout vectors laying in their corresponding readout zones increases. This might explain the discrepancy between the ``networks'' in Fig.~\ref{fig4} and our network (Fig.~\ref{fig2}). We sampled $100$ random readouts from a representational space spanning $1$ to $50$ dimensions, and asked how many of those lay in their corresponding readout zones, similar to the analysis above. We did the same for the readout matrix of our trained RNN - we sampled $1$ to $50$ dimensions of the representational space. We repeated both analysis $100$ times to get confidence bounds. As seen in Fig.~\ref{fig5}, all the $100$ readout vectors lay in their corresponding readout zones once the dimensionality of the representational space was around $30$ (much lower than the dimensionality of the representational space of our RNNs which is $512$), for both the random and trained readouts. The proportion of readout vectors being in their corresponding zones increases faster for the random readouts as they probably span a larger subspace than the trained readouts (owing to correlations across classes), leading to a lower proportion of closely located readout vectors.

In summary, in our RNNs, indeed the readout vectors lie in readout zones which are high-dimensional conical structures anchored to origin, as schematized in Fig.~\ref{fig3}B. 

Do these insights generalize to RNNs with readout biases? Adding bias did not affect the classification accuracy: both, the  RNN from Fig.~\ref{fig2} and the same version trained with readout bias, had a maximum accuracy of ~$53\%$ on the ``testplus'' split of MiniEcoset. Removing the bias after training with it also did not change the maximum accuracy. Relatedly, in the RNN trained with readout bias, the norm of the bias was $0.27$ whereas the norms of the representations lay in the range [$5.1$, $17.9$] with a mean of $8.1$. These results suggest that the regime in which our RNNs with readout bias operate would lead to readout zones similar to the ones seen in Fig.~\ref{fig4} (bottom), which makes our representational arrangement insights (Fig.~\ref{fig3}B) generalize to RNNs with readout biases.

\subsection{Representational norms in feedforward networks predict $t_{stable}$ in the RNN}
\label{ssc:ffnet}

We asked whether the arrangement of norms found in the RNN, shown in Fig.~\ref{fig2}B, could be due to the nature of the computations in the feedforward sweep: images that take longer to be classified by the RNN are the ones that are not ``prototypical'' of their category. This would be reflected as them having lower alignment with the feedforward weights (feature kernels), leading to lower neural responses i.e. norms. To test this idea, we assessed if the arrangement of norms found in the feedforward sweep of the RNN (``BNet''; trained without recurrence on MiniEcoset) predict the time taken by the RNN to stably classify the images. We also assessed if this arrangement of norms is independent of network architecture by testing the norms from a ResNet18~\cite{he2016deep} trained on MiniEcoset.

\begin{figure}[t]
  \begin{center}
  \includegraphics[width=\textwidth]{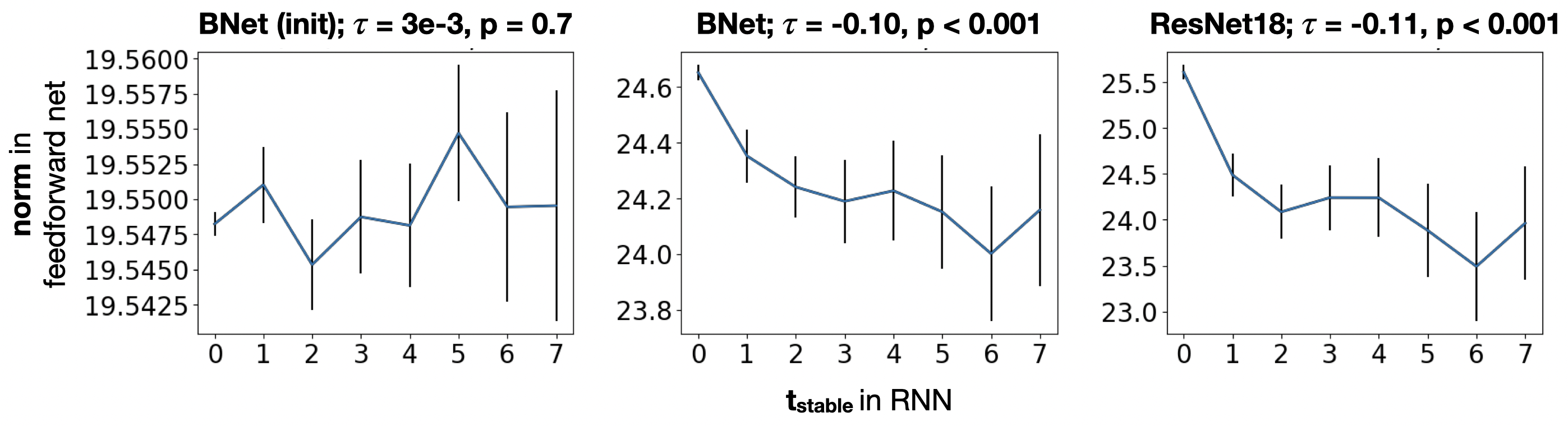}
  \end{center}
  \caption{The norms of the feedforward networks' pre-readout responses to the stably classified images, considered in Fig.~\ref{fig2}, are related to their $t_{stable}$ values from the RNN. A negative correlation is found for the trained feedforward networks, in line with previous results: higher norms are associated with quicker stable classification.}\label{fig6} 
\end{figure}

For each of the stably classified images considered in Fig.~\ref{fig2}, we extracted the norms of the responses at the pre-readout layers for each network, and related them to the corresponding $t_{stable}$ values from the RNN. As seen in Fig.~\ref{fig6}, a negative relationship was found: representations with higher norms in the feedforward networks corresponded to images which took longer to be stably classified by the RNN. Crucially, this relationship was not found in a randomly initialised BNet, indicating that training on the dataset is essential to acquire this norm arrangement. 

This result suggest that while the representational arrangement found in the RNNs dovetails with the constraint of ``equal movement across representations'' (Fig.~\ref{fig2}A), the arrangement might not have emerged due to the constraint. Rather the arrangement is a property of the feedforward network, which is then used by the recurrent computations to move these non-prototypical representations into their correct class.

\subsection{Stability analysis}
\label{ssc:stability}

The overall decrease over time in the change in representations seen in Fig.~\ref{fig2}A suggests that these RNNs might have convergent dynamics. However, to establish convergent dynamics, the change in representations needs to be assessed past the timesteps the RNNs were trained for. Each of the $4$ RNNs (analysed in Fig.~\ref{fig3}), which were trained with classification losses aggregated over $10$ timesteps, were unrolled to $80$ timesteps to assess their dynamics beyond their training. If the RNNs indeed have convergent dynamics i.e. are stable, the change in representations should decrease with timesteps and be much smaller than the norm of the representations. Additionally, the classification accuracies should not drop in the timesteps beyond the training timesteps. We monitored the relative change in the representation $R$ at timestep $t$: $log_{10}(\left \| R_{t}-R_{t-1} \right \|) - log_{10}(\left \|R_{t-1}\right \|)$), and the classification accuracy on the test set of MiniEcoset, across the $80$ timesteps.

\begin{figure}[t]
  \begin{center}
  \includegraphics[width=\textwidth]{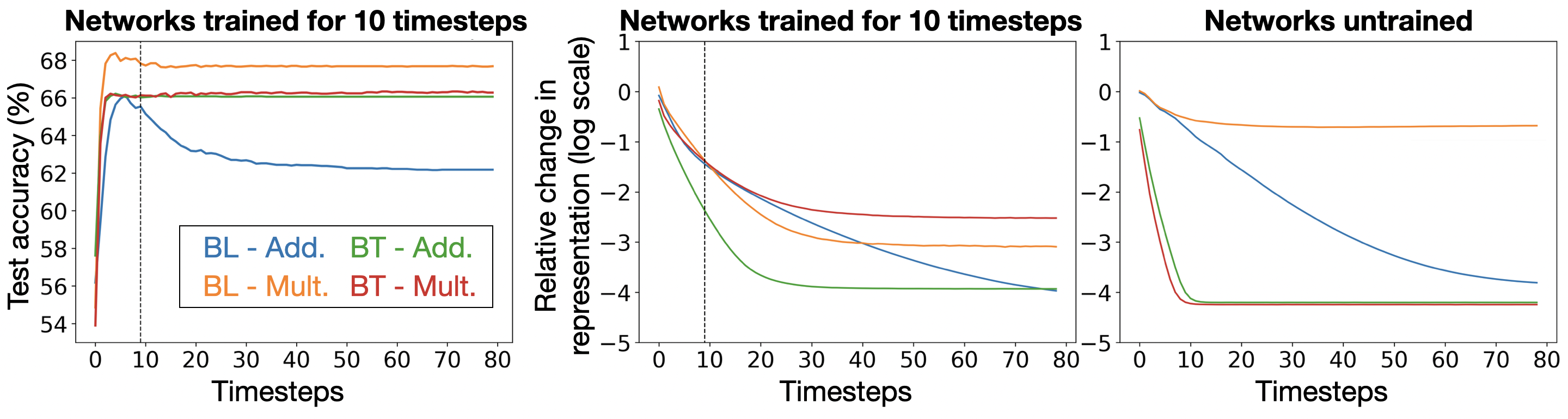}
  \end{center}
  \caption{Assessing RNN stability. The RNNs were trained over $10$ timesteps and were unrolled to $80$ timesteps post-training. (Left) The classification performance of the $4$ RNNs, analysed in Fig.~\ref{fig3}, did not drop substantially past the point they were trained for. (Middle) The relative change in representations decreased through time and was orders of magnitude smaller than the norms of those representations. These results suggest these RNNs have convergent, stable, dynamics. (Right) These stable dynamics existed pre-training.}\label{fig7} 
\end{figure}

As seen in Fig.~\ref{fig7}, the relative change in the representations (averaged across the test set) decreased with timesteps and was at least $2$ orders of magnitude smaller than the norms of those representations. This shrinking change in representations was accompanied by a sustained classification performance (the worst performance loss was $\sim6\%$ of the performance at timestep $10$). Additonally, this signature of decreasing relative change in representations was also found in these RNNs before training.

These results suggest that these RNNs are indeed exhibiting convergent dynamics i.e. they are stable. Note that previous research indicated that RNNs trained with backpropagation through time (BPTT) do not show such convergent dynamics beyond the training timesteps~\cite{linsley2020stable}. Future research would address these discrepancies in the dynamics, by assessing the points of divergence between the RNNs used in previous research and the current class of RNNs, in terms of the architecture, the dataset, the classification objective, and the learning rule.

\end{document}